\documentclass[runningheads]{llncs}

\usepackage{wrapfig}
\usepackage{float}
\usepackage{eccv}

\usepackage{eccvabbrv}
\usepackage{mathtools}
\usepackage{graphicx}
\usepackage{booktabs}
\usepackage{multirow}
\usepackage[table]{xcolor}
\usepackage[accsupp]{axessibility}  
\usepackage{booktabs}
\usepackage{subcaption}
\usepackage{makecell}
\usepackage{marvosym}

\usepackage[pagebackref,breaklinks,colorlinks,citecolor=eccvblue]{hyperref}

\usepackage{orcidlink}
\usepackage{cite}
\usepackage{pifont}

\providecommand{\thetitle}{}
\let\oldtitle\title
\renewcommand{\title}[1]{\oldtitle{#1}\renewcommand{\thetitle}{#1}}
\newcommand{\maketitlesupplementary}{
    \newpage
    \begin{center}
        \Large
        \textbf{\thetitle}\\[0.5em] 
        Supplementary Material\\[1.0em]
    \end{center}
}

\begin{document}

\title{Online Segment 3D Gaussians via \\ Launching Virtual Drones} 

\titlerunning{Online Segment 3D Gaussians via Launching Virtual Drones}

\author{Liwei Liao\inst{1,2,3} \and
Rongjie Wang\inst{2} \and
Ronggang Wang\inst{1,2,3}\Letter}

\authorrunning{L.~Liao et al.}

\institute{Guangdong Provincial Key Laboratory of Ultra High Definition Immersive Media Technology, Shenzhen Graduate School, Peking University, Shenzhen, China \and
Pengcheng Laboratory, Shenzhen, China \and
Peking University, Beijing, China\\
\email{levio@pku.edu.cn, rgwang@pkusz.edu.cn}}

\maketitle

\begin{abstract}
Interactive segmentation of 3D Gaussians offers a compelling opportunity for real-time manipulation of 3D scenes, thanks to the real-time rendering capability of 3D Gaussian Splatting (3DGS). However, existing methods require a time-consuming per-scene setup—typically tens of seconds or even minutes—before interactive segmentation can begin on a raw 3DGS scene. This setup involves multi-view mask preparation, mask lifting, and feature distillation, creating a major bottleneck for online applications.
To address this limitation, we aim to completely eliminate the setup stage for interactive 3DGS segmentation while keeping the segmentation time practical (under 1 second). In this work, we present \textbf{SAGO} (\underline{S}egment \underline{A}ny \underline{G}aussians \underline{O}nline), a novel setup-free framework for interactive 3DGS segmentation. By introducing \textbf{virtual drones}, our method reframes the 3D segmentation problem as an online Next-Best-View (NBV) planning task formulated within a Markov process. Extensive experiments demonstrate that \textbf{SAGO} can extract clean 3D assets directly from 3D Gaussians with sub-second latency, thereby enabling a broad range of downstream applications such as object manipulation and scene editing. Moreover, our method achieves over a \textbf{50}$\times$ speedup compared to the previous setup-free 3DGS segmentation frameworks.

\keywords{Interactive 3D Segmentation \and 3D Gaussian Splatting \and Next Best View \and Online Planning \and 3D Segmentation \and Active Vision}

\end{abstract}

\section{Introduction}
\label{sec:intro}
\begin{figure*}[t]
  \centering
    \setlength{\abovecaptionskip}{0.3cm}
    \setlength{\belowcaptionskip}{-0.2cm}
  \includegraphics[width=1\linewidth]{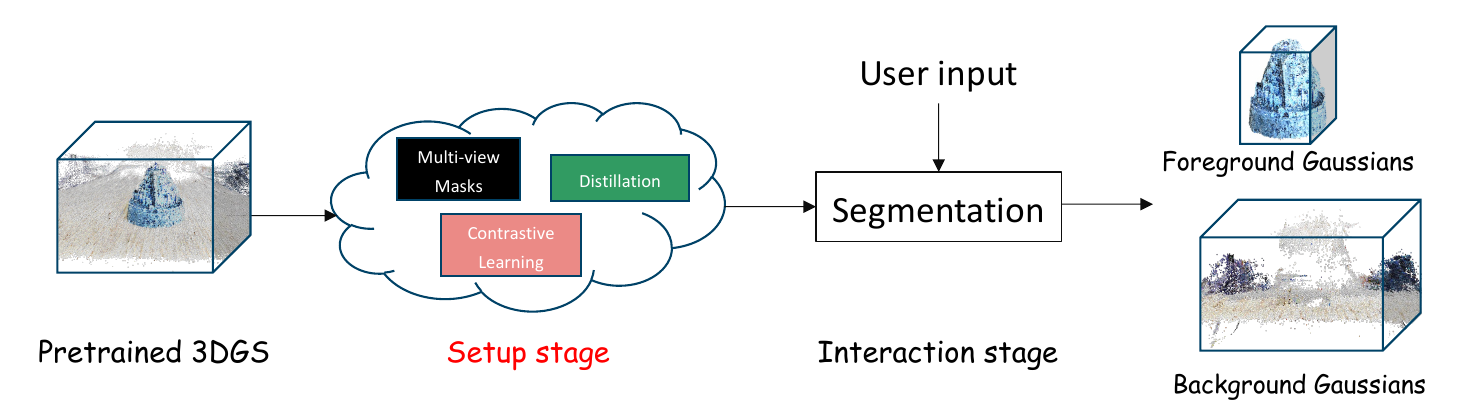}
  \caption{\textbf{The common pipeline of interactive 3DGS segmentation.} Existing methods require a setup stage of tens of seconds or even minutes before interactive segmentation can be performed on a given raw 3DGS scene.}
  \label{fig:intro}
\end{figure*}

In recent years, 3D Gaussian Splatting (3DGS)~\cite{3DGS} has seen remarkable progress in 3D representation and has shown great potential as a medium for 3D scene understanding in downstream applications such as embodied AI~\cite{duan2022survey,liu2025building}, robotics~\cite{zhu20243d, lu2024manigaussian}, data synthesis~\cite{li2024robogsim, xu2025nvpose} and immersive media~\cite{wu20244dgs,yan2024saro, wu2025swift4d, wu2025localdygs, Liang_2026_CVPR, yang2026i3dv, liang2024high, peng2024structure, deng2026pano, wang2026sap, xiong2026intrinsic}. To equip 3DGS with capabilities of interaction, editing, and scene understanding, a series of 3D segmentation methods~\cite{GauGroup, cen2025segment,liao2026gaussiantrimmer, SAGD, shen2024flashsplat, jain2024gaussiancut, choi2024click, zhang2025cob, qin2024langsplat, wu2024opengaussian, liao2024clipgs,liao2026zero, zheng2024surface, zhao2025isegman, zheng2025space, li2025instancegaussian} have achieved remarkable progress. Most of them follow a pipeline as shown in~\cref{fig:intro} that requires a scene-specific setup stage to pre-process the pretrained 3DGS scene to make it interactively segmentable. The setup stage typically involves pre-segmenting training views with 2D segmentation models, such as the Segment Anything (SAM) series~\cite{kirillov2023segment, ravi2024sam2, carion2025sam3}, to generate 2D masks. These masks are subsequently utilized for mask-lifting, feature distillation, or other forms of optimization. Due to the setup stage, existing methods require an additional time of tens of seconds or even minutes to process a new 3DGS scene before interactive segmentation can be performed, which significantly limits their efficiency and user experience in practical applications.

To eliminate the setup stage, we explore an online and inference-only approach for 3DGS segmentation. Our goal is to enable interactive segmentation of a given raw 3DGS scene within a user-tolerable time frame after the user provides a prompt describing the target object (e.g., clicks, boxes, or text), while achieving comparable segmentation quality to existing methods requiring a per-scene setup stage (i.e., offline methods).
To achieve this, we revisit the problem of online view planning within the context of active robotic vision~\cite{zeng2020view, zhang2025novel, jin2024gs}, where autonomous drones are deployed to explore unknown regions of physical scenes for high-fidelity digital reconstruction. Each drone is equipped with a camera that can be maneuvered to capture images from various viewpoints, and the Next-Best-View (NBV) planning module is responsible for determining the optimal camera parameters to maximize the coverage of the target object. This NBV planning problem shares a significant similarity with our 3D segmentation task: \textit{each additional effective 2D view can provide substantial cues for the final 3D task}. In our 3D segmentation framework, 2D segmentation masks from novel viewpoints serve as spatial constraints to effectively ``prune'' background Gaussians that project outside the mask contours. 

Building on this insight, we introduce \textbf{virtual drones} and formulate the 3D segmentation as an NBV planning problem for virtual drones. Starting from the initial view (i.e., the view for user interaction), the virtual drone iteratively plans its NBV based on the current state in a Markov Process framework. At each step, the virtual drone captures a new view of the scene, and the 2D mask obtained from SAM is used to update the state by pruning background Gaussians that project outside the mask contours. By iteratively updating the state and planning the NBV, our method can effectively segment the target object in 3D without requiring a time-consuming setup stage, while still maintaining high segmentation quality.
In summary, our key contributions are as follows:
\begin{itemize}

\item We propose \textbf{SAGO}, an effective and efficient online 3DGS segmentation framework. SAGO completely eliminates the time-consuming scene-specific setup stage required by most existing methods, while delivering segmentation performance comparable (or superior) to state-of-the-art offline approaches.

\item We introduce the concept of \textbf{virtual drones} and present a principled formulation that recasts 3D online segmentation as a Next-Best-View (NBV) planning problem for a virtual drone navigating and observing the scene.

\item In contrast to previous setup-free methods, ours achieves over \textbf{50$\times$} inference speedup while simultaneously delivering superior segmentation quality.

\end{itemize}

\section{Related Works}
\noindent\textbf{Optimization-based 3DGS Segmentation.} 
Most existing approaches~\cite{GauGroup, cen2025segment, choi2024click, zhang2025cob, qin2024langsplat, zhu2025rethinking} rely on a preparatory setup stage in which 2D segmentation masks are first generated for a set of training views, followed by a dedicated optimization procedure—commonly involving mask lifting, feature distillation, or contrastive learning. 
For example, SAGA~\cite{cen2025segment} employs scale-aware contrastive learning to extract scale-gated affinity features, while Click-Gaussian~\cite{choi2024click} aggregates global feature candidates to provide consistent supervision for Gaussian feature learning. 
COB-GS~\cite{zhang2025cob} introduces boundary-adaptive Gaussian splitting to better resolve ambiguous regions near object boundaries. 
LangSplat~\cite{qin2024langsplat} distills semantic knowledge from CLIP~\cite{radford2021learning} into 3D Gaussians using an auto-encoder architecture. 
These optimization-based methods typically require iterative gradient-based optimization, resulting in substantial computational time and memory costs during the setup phase.

\noindent\textbf{Online 3DGS Segmentation}. 
Recent methods~\cite{jain2024gaussiancut, zhao2025isegman, SAGD, shen2025trace3d, sun2025sagonline,liao2026zero} have explored optimization-free approaches to 3D Gaussian Splatting (3DGS) segmentation, a paradigm we refer to as \emph{online 3DGS segmentation}. 
These methods generally bypass time-consuming per-scene optimization and gradient-based backpropagation. 
For example, GaussianCut~\cite{jain2024gaussiancut} constructs a graph over Gaussians and reformulates 3D segmentation as a graph partitioning problem. 
iSegMan~\cite{zhao2025isegman} introduces epipolar-guided interaction propagation combined with visibility-based Gaussian voting to achieve fully training-free segmentation. 
SAGD~\cite{SAGD} employs Gaussian decomposition to alleviate jagged boundary artifacts. 
Trace3D~\cite{shen2025trace3d} proposes Gaussian Instance Tracing (GIT), which explicitly associates each Gaussian with its corresponding instance labels across multiple views through a learned weight matrix.

Although these online 3DGS segmentation methods eliminate per-scene optimization, most still rely on non-trivial preprocessing steps and are not truly free of any setup phase. 
More critically, their per-interaction latency remains significantly higher than that of optimization-based approaches, which benefit from precomputed features. 
For example, GaussianCut~\cite{jain2024gaussiancut} performs online graph construction, while iSegMan~\cite{zhao2025isegman} requires online epipolar-guided interaction propagation and visibility-based Gaussian voting—both of which incur considerable computational cost during user interactions. 
In contrast, our proposed method introduces the virtual drones and leverages the NBV planning theory to achieve a completely setup-free pipeline with highly efficient interaction times, typically under 1 second per interaction.


\section{Method}
We propose SAGO, an online 3DGS segmentation method based on virtual drone NBV planning, enabling interactive segmentation without the need for a time-consuming setup stage. In this section, we first introduce the relevant knowledge of 3D Gaussian Splatting (3DGS), particularly the geometric projection of 3D Gaussians onto the 2D plane, which allows us to use 2D masks for background filtering, forming the theoretical basis for our online segmentation method. Next, we will detail the state initialization process and the online view planning and state update mechanism, which are core components of our method.

\begin{figure*}[t]
  \centering
    \setlength{\abovecaptionskip}{0.3cm}
    \setlength{\belowcaptionskip}{-0.2cm}
  \includegraphics[width=1\linewidth]{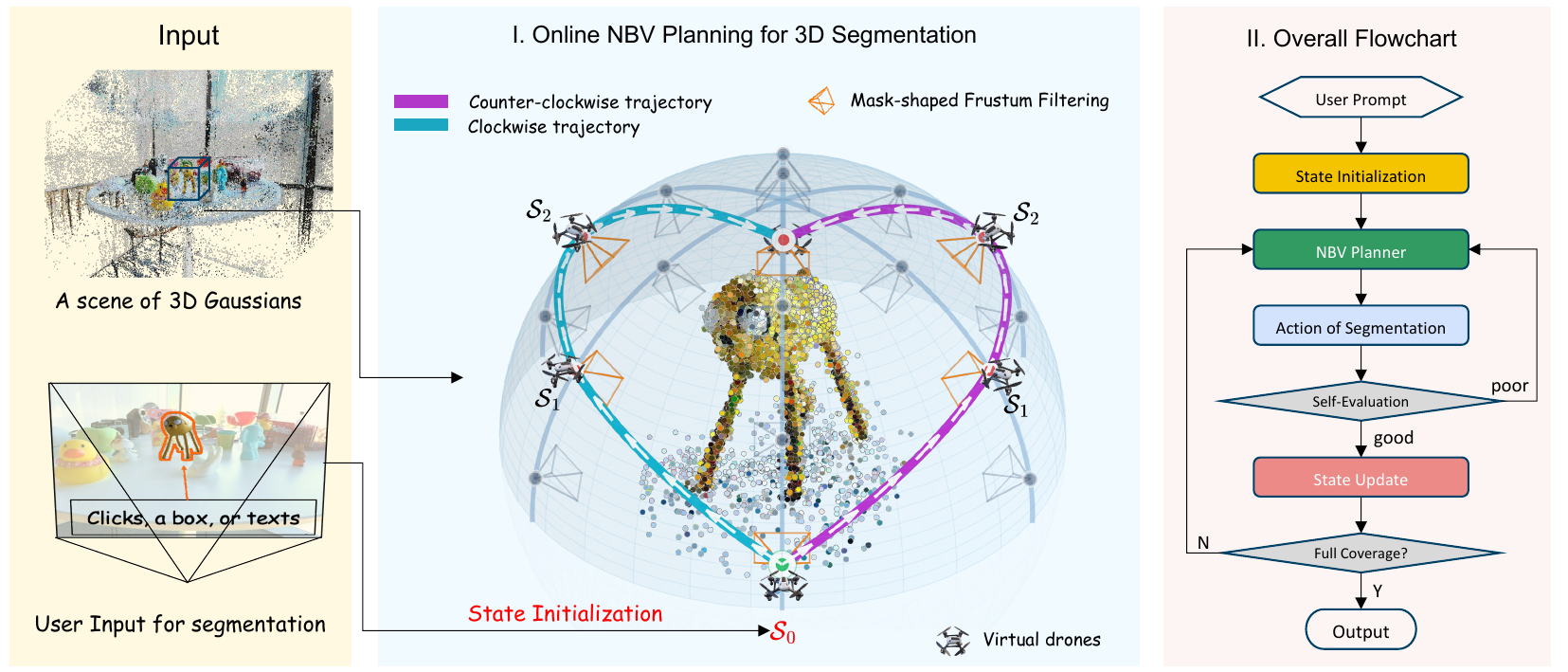}
  \caption{\textbf{The pipeline of SAGO.} 
  Following user input, SAGO launches virtual drones in both clockwise and counter-clockwise directions from the interaction viewpoint to explore Next-Best-Views. Each newly captured view provides a segmentation mask that effectively filters out background Gaussians via Mask-shaped Frustum Filtering (MFF), with the process repeating until full coverage is achieved. 
  }
  \label{fig:framework}
\end{figure*}

\subsection{Preliminary: 3D Gaussian Splatting}
3D Gaussian Splatting (3DGS)~\cite{3DGS} has emerged as a prominent approach for real-time novel view synthesis, representing scenes using a collection of explicit 3D Gaussians, achieving an excellent trade-off between rendering quality and speed. A set of pretrained 3D Gaussians $\mathcal{G} = \{G_i\}_{i=1}^N$ is obtained by optimizing the parameters of the Gaussians through differentiable rendering techniques. Each Gaussian $G_i$ is parameterized as $G_i=\{x_i,q_i,s_i,c_i, o_i\}$, where $x_i\in \mathbb{R}^3$ is the 3D position, $q_i\in \mathbb{R}^4$ is the quaternion representing the rotation, $s_i\in \mathbb{R}^3$ is the scale, $c_i\in \mathbb{R}^{48}$ is the view-dependent color represented by three-order spherical harmonics~\cite{fridovich2022plenoxels}, and $o_i\in \mathbb{R}$ is the opacity. Based on their role in 3D segmentation, Gaussian attributes can be categorized into two groups: \textbf{geometric attributes} (position, covariance (including rotation and scaling)) and \textbf{appearance attributes} (color and opacity). Geometric attributes can serve as a discriminative basis for distinguishing the foreground from the background, while appearance attributes provide extra semantic information to distinguish different instances. In the rendering stage, the 2D covariance matrix $\Sigma_i^{2D}$ of a Gaussian $G_i$ can be computed as follows:
\begin{equation}
\Sigma_i^{2D}=J W (R_iS_iS_i^TR_i^T) W^T J^T,
\end{equation}
where $R_i$ is the rotation matrix derived from the quaternion $q_i$, $S_i$ is the diagonal matrix of the scale $s_i$, $W$ is the projection matrix, and $J$ is the Jacobian matrix of the projection. The 2D covariance $\Sigma_i^{2D}$ represents the $i$-th Gaussian's shape in the 2D plane, and the projected center of the Gaussian is given by $x_i^{2D} = project(x_i)$. Thus, the $i$-th Gaussian can be projected onto the 2D plane as an ellipse (referred to as a "splat"), defined by  $\{x_i^{2D}, \Sigma_i^{2D}\}$. This projection allows us to leverage 2D masks to inform the segmentation of the 3D Gaussians (see Fig.~\ref{fig:mask-shaped}), forming the basis for our online segmentation pipeline.

\subsection{Overview}
$\textbf{Problem Definition.}$ Given a set of 3D Gaussians $\mathcal{G}$ representing a scene and a user-provided prompt (e.g., point clicks, scribbles, or text) on a 2D view, interactive 3D segmentation aims to partition $\mathcal{G}$ into two disjoint, non-empty subsets $\mathcal{A}$ and $\mathcal{B}$. Here, $\mathcal{A}$ represents the Gaussians corresponding to the target object (the foreground), while $\mathcal{B}$ represents the background, such that $\mathcal{A} \cup \mathcal{B} = \mathcal{G}$ and $\mathcal{A} \cap \mathcal{B} = \emptyset$. This can be formulated as:
\begin{equation}
\mathcal{A}, \mathcal{B} = \operatorname{segment}(\mathcal{G}, \texttt{prompt}).
\end{equation}

\noindent$\textbf{Overall pipeline}.$ As shown in~\cref{fig:framework}, we model the interactive 3D segmentation process as an NBV planning problem for virtual drones, where the state at each time step $t$ is represented as $S_t$. The virtual drones explore the pre-defined search space to find optimal views for segmentation. At each new view, a segmentation mask is generated via SAM, which is then used to filter out background Gaussians. 
With the updated segmentation results, the state is updated and the next NBV is planned, iterating until full coverage of the target object is achieved. The core components of this pipeline include state initialization, NBV planning, and state update, which will be detailed in the following sections.

The key notations used in this paper are summarized in~\cref{tab:notation} for clarity and reference throughout the subsequent sections.

\subsection{State Initialization}
We leverage virtual drone Next-Best-View (NBV) planning to address the online 3DGS segmentation problem, which can be modeled as a Markov process where the next state depends only on the current state. Starting from an initial 2D user-prompted view, which serves as the virtual drone's origin, we iteratively plan a sequence of NBVs. At each new viewpoint, a segmentation mask is generated via SAM with memories saved from previous states to filter out background Gaussians, continuing until full coverage is achieved. We define the state at each step using a four-tuple of variables: the current set of foreground Gaussians $\mathcal{A}^{(t)}$, the virtual drone’s pose $\mathbf{v}_t$, a memory bank $\mathcal{M}_t$ for SAM tracking, and a coverage indicator $\tau_t$. The drone’s pose is represented in an object-centric spherical coordinate system, consisting of the target center $\mathbf{c}_t$, radius $r_t$, yaw $\phi_t$, and pitch $\theta_t$. This can be formulated as:
\begin{equation}
\begin{split}
\mathcal{S}_t &= \{\mathcal{A}^{(t)}, \mathbf{v}_t, \mathcal{M}_t, \tau_t\}, \\
\mathbf{v}_t &= \{\mathbf{c}_t, r_t, \phi_t, \theta_t\}.
\end{split}
\label{eq:state}
\end{equation}

\begin{figure}[t]
\centering
\begin{minipage}{0.6\textwidth}
\centering
\small
\setlength{\tabcolsep}{2mm}
\renewcommand{\arraystretch}{1.3}
\captionof{table}{\textbf{A list of key notations used in this paper.}}
\label{tab:notation}
\resizebox{\linewidth}{!}{%
\begin{tabular}{cl}
\noalign{\hrule height 1.2pt}
\textbf{Symbol} & \textbf{Description at Step $t$} \\
\midrule
$\mathcal{A}^{(t)}, \mathcal{B}^{(t)}$ & Foreground/background subsets at step $t$ \\
$S_t$ & Segmentation state at step $t$ \\
$\mathcal{M}_t$ & Segmentation memory bank at step $t$ \\
$\tau_t$ & Coverage indicator at step $t$ \\
$\mathbf{v}_t$ & Camera pose of virtual drone at step $t$ \\
\hline
$\mathbf{c}_t$ & Target center/the point virtual drones look at\\
$r_t$ & The distance between $\mathbf{c}_t$ and the virtual drone\\
$\phi_t$ & The yaw angle of the virtual drone at step $t$\\
$\theta_t$ & The pitch angle of the virtual drone at step $t$\\
\noalign{\hrule height 1.2pt}
\end{tabular}%
}
\end{minipage}%
\hfill
\begin{minipage}{0.36\textwidth}
\centering
\includegraphics[width=\linewidth]{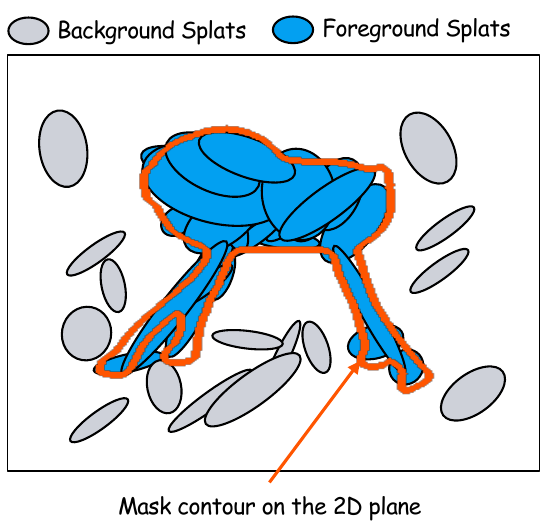}
\captionof{figure}{Mask-shaped Frustum Filtering (MFF) illustration.}
\label{fig:mask-shaped}
\end{minipage}
\end{figure}



According to~\cref{eq:state}, we can initialize the state $\mathcal{S}_0 = \{\mathcal{A}^{(0)}, \mathbf{v}_0, \mathcal{M}_0, \tau_0\}$. The initial virtual drone pose $\mathbf{v}_0$ is derived from the camera pose of the interaction viewpoint with yaw $\phi_0$ set to $0$, and $\tau_0$ is set to 0 initially. The memory bank $\mathcal{M}_0$ is initialized via a SAM memory encoder~\cite{ravi2024sam2} based on the user's prompt. The initial foreground Gaussian set $\mathcal{A}^{(0)}$ is a subset of $\mathcal{G}$, with a part of background Gaussians pruned by mask-shaped frustum filtering based on the initial mask.

\noindent\textbf{Mask-shaped Frustum Filtering (MFF).} The 2D mask on each view can be used to filter out certain background Gaussians, and we call this process mask-shaped frustum filtering. Specifically, as shown in~\cref{fig:mask-shaped}, we project each Gaussian onto the 2D plane and check its spatial relationship with the mask region. If the Gaussian is determined to overlap with the mask, we consider it as a potential foreground candidate and include it in $\mathcal{A}$. Otherwise, we classify it as background set $\mathcal{B}$ and exclude it from the foreground set. This filtering process helps to reduce the number of Gaussians that need to be considered in subsequent iterations, improving the efficiency of the online segmentation process. The MFF in each view can exclude a group of Gaussians $\Delta \mathcal{A}$ from the foreground set $\mathcal{A}$ to the background set $\mathcal{B}$, which can be formulated as:
\begin{equation}
  \begin{split}
\mathcal{A}^{(t+1)} = \mathcal{A}^{(t)} - \Delta \mathcal{A}^{(t)}, \\
\mathcal{B}^{(t+1)} = \mathcal{B}^{(t)} + \Delta \mathcal{A}^{(t)}.
\end{split}
\end{equation}

MFF can be implemented in two ways: center-based MFF and splat-based MFF. Center-based MFF directly projects the center of each Gaussian onto the 2D mask, classifying it as background if the center falls outside the mask. Splat-based MFF, on the other hand, projects each Gaussian as a splat and classifies it as foreground if there is any overlap between the splat and the mask; otherwise, it is classified as background. We adopt the center-based MFF in our implementation for its simplicity and efficiency, and the performance is even better (see~\cref{sec:ablation}).

\subsection{Online NBV Planning and State Update}
\label{sec:NBV}
\begin{figure*}[t]
  \centering
    \setlength{\abovecaptionskip}{0.1cm}
  \includegraphics[width=1\linewidth]{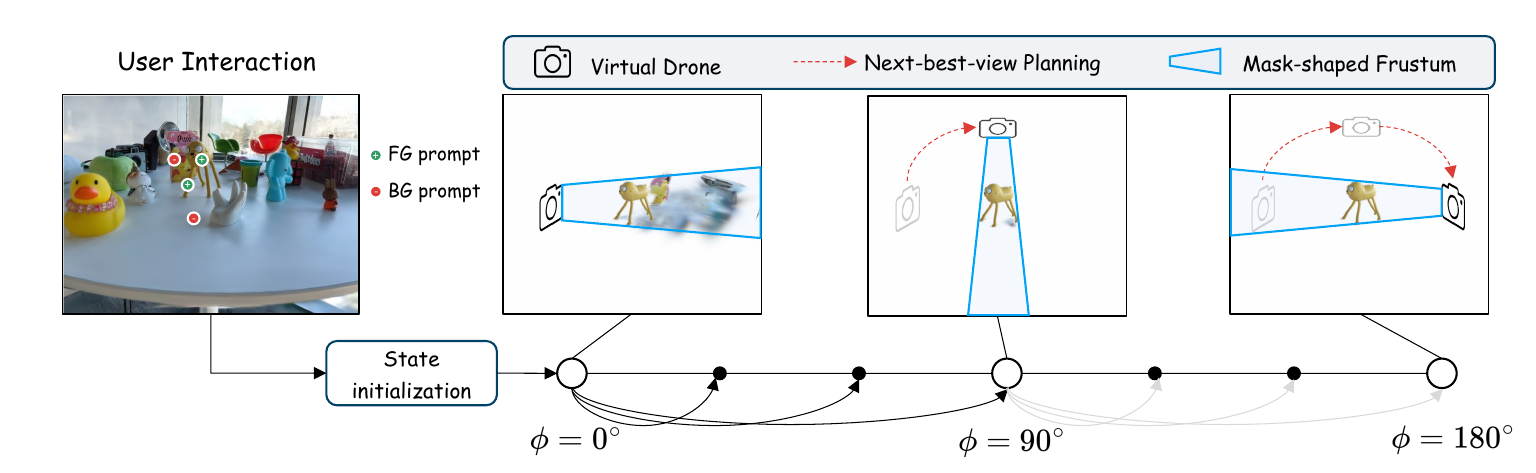}
  \caption{The illustration of NBV-based online 3D segmentation (clockwise direction). 
  }
  \label{fig:state_update}
\end{figure*}

As illustrated in~\cref{fig:framework}, we launch two virtual drones simultaneously from the interaction viewpoint, exploring NBVs in both clockwise and counter-clockwise directions. Each drone is responsible for exploring a 180-degree yaw range. Due to the symmetry of the two trajectories, we demonstrate the update process of one trajectory, which will be explained in the following using the clockwise direction as an example (see~\cref{fig:state_update}).

To reach a full coverage of the target object, we leverage an object-centric spherical coordinate system for NBVs planning. The camera pose of the virtual drone at step $t$ is represented as $\mathbf{v}_t = (\mathbf{c}_t, r_t, \phi_t, \theta_t)$, where $\mathbf{c}_t$ is the target center that the virtual drone looks at, $r_t$ is the distance between $\mathbf{c}_t$ and the virtual drone, $\phi_t$ is the yaw angle of the virtual drone at step $t$, and $\theta_t$ is the pitch angle. The target center $\mathbf{c}_t$ is initialized as the average of the 3D positions of the Gaussians that are projected inside the initial mask $m_0$ at step 0, and is updated as the average of the 3D positions of the currently segmented foreground Gaussians $\mathcal{A}^{(t-1)}$ at step $t$ (if $t \geq 2$), as follows:
\begin{equation}
\mathbf{c}_t = 
\begin{dcases} 
\frac{1}{|X_{\text{reproj}(m_0)}|} \sum_{x \in X_{\text{reproj}(m_0)}} x, & \text{if } t = 0,1, \\
\frac{1}{|X_{\mathcal{A}^{(t-1)}}|} \sum_{x \in X_{\mathcal{A}^{(t-1)}}} x, & \text{if } t \geq 2,
\end{dcases}
\label{eq:center}
\end{equation}
where $X$ denotes the 3D positions of Gaussians or 3D points.
\cref{eq:center} actually determines the geometric centroid or 3D location of the target object, and the centroid $\mathbf{c}_t$ acts as the origin of the spherical coordinate system for NBV planning. In the initial stage, our method can estimate the centroid $\mathbf{c}_t$ roughly through the 3D points obtained by depth back-projection of the 2D mask from the initial view, which usually has a certain offset and lies near the surface of the object. The offset always leads to the object being not well centered in the next viewpoint when the yaw span is too large. 
But when the foreground set $\mathcal{A}$ is updated closer to the ground truth, the centroid $\mathbf{c}_t$ can be updated to be more accurate using the mean of the 3D positions of the current $\mathcal{A}^{(t)}$. 

The radius $r_t$ is initialized to a fixed value that is the distance between the target center $\mathbf{c}_t$ and the start point of the virtual drones, and remains constant during the state update process to make the virtual drone maintain a consistent distance from the target object.

The yaw angle $\phi_t$ controls the horizontal circular motion of the virtual drone, starting at an initial value of $0^{\circ}$. When $\phi_t$ reaches $360^{\circ}$, the virtual drone completes a full revolution around the object and returns to its starting position. In our setup, the clockwise direction is defined as $\phi$ increasing from $0^{\circ}$ to $180^{\circ}$.

The pitch angle $\theta_t$ governs the vertical movement of the virtual drone, enabling it to explore different elevations around the target object and thereby achieve more comprehensive segmentation coverage. This angle is constrained to the interval $[0^{\circ}, 90^{\circ}]$.

Since $r_t$ remains constant and $\mathbf{c}_t$ is deterministically computed from the current foreground set, our NBV planning primarily focuses on determining the optimal yaw angle $\phi_t$ and pitch angle $\theta_t$ at each time step. We employ an \textbf{Exploration-Evaluation (EE)} strategy to determine the next angles $\phi_{t+1}$ and $\theta_{t+1}$ for the virtual drone, and set the search space for NBV planning as candidate sets $\Phi=\{90^\circ, 60^\circ, 30^\circ\}$ and $\Theta=\{\theta_0, 0^{\circ}, 30^{\circ}, 60^{\circ}\}$ for yaw and pitch angles, respectively. At time step $t$, the virtual drone searches over candidate combinations of $\Delta\phi \in \Phi$ and $\theta \in \Theta$ to select the optimal pose  as the next-best view (NBV) $\mathbf{v}_{t+1}$ and navigates to this position. We use $\mathbf{v}=\{\mathbf{c}_t, r_t, \phi_t + \Delta \phi, \theta\}$ to denote the candidate pose that is selected as the NBV, which can be formulated as:
\begin{equation}
\begin{aligned}
\mathbf{v}^* = \arg\max_{\mathbf{v}_{(\Delta \phi,\theta)}}  
\bigl|\Delta \mathcal{A}'(\mathbf{v})\bigr|
\quad \text{s.t.} \quad \mathrm{mIoU}\bigl({I}(\mathbf{v}_0, \mathcal{A}^{(t)} - \Delta \mathcal{A}'(\mathbf{v})), m_0\bigr) \geq \sigma,
\end{aligned}
\label{eq:nbv_planning}
\end{equation}
where

\begin{equation}
\Delta \mathcal{A}' = \texttt{MFF}(\texttt{SAM}(\mathcal{I}(\mathbf{v}, \mathcal{A}^{(t)}), \mathcal{M}_t), \mathcal{A}^{(t)}).
\label{eq:action_of_segmentation}
\end{equation}

\cref{eq:action_of_segmentation} denotes the action of segmentation that finds the change in the foreground Gaussian set $\mathcal{A}$ after performing Mask-shaped Frustum Filtering (MFF) based on the segmentation mask obtained from SAM with memory $\mathcal{M}_t$ at the candidate view $\mathcal{I}(\mathbf{v}, \mathcal{A}^{(t)}))$. \cref{eq:nbv_planning} formulates the NBV planning as an optimization problem that aims to find the candidate view $\mathbf{v}$ that maximizes the number of Gaussians that can be filtered out as background (i.e., $|\Delta \mathcal{A}'|$) while ensuring that the segmentation result of the candidate view maintains a certain level of consistency with the initial mask $m_0$ as measured by mIoU ($\sigma$ denotes the mIOU threshold), which helps to avoid cascading over-segmentation errors along the Markov chain. As illustrated in~\cref{fig:state_update}, starting from $\phi = 0^{\circ}$, the virtual drone first attempts a large yaw increment of $90^{\circ}$. If SAM successfully tracks the target, it flies directly to $\phi = 90^{\circ}$ for state update. Otherwise, it falls back to a smaller increment of $60^{\circ}$ and finally to $30^{\circ}$. In the case of successful tracking at each step, the virtual drone can achieve full coverage with only two exploration steps, which makes our NBV-based online segmentation efficient.

\textbf{State Update}. After the NBV $\mathbf{v}_{t+1}$ is selected ($\mathbf{v}_{t+1} = \mathbf{v}^*$), the virtual drone navigates to this new position and captures a new view of the scene. The segmentation mask obtained from SAM at this new viewpoint is then used to update the foreground Gaussian set $\mathcal{A}$ and the background Gaussian set $\mathcal{B}$ through MFF. This process effectively refines the segmentation by filtering out additional background Gaussians based on the new view, and updates the state of the system for the next iteration of NBV planning. We refer to this iterative process of updating the segmentation state based on new views as \textbf{State Update}, which is crucial for achieving full coverage of the target object through successive NBV explorations. The core variables in state $\mathcal{S}_{t+1}$ can be updated as follows:

\begin{equation}
\mathcal{A}^{(t+1)} = \mathcal{A}^{(t)} - \Delta \mathcal{A}(\mathbf{v}^*), \quad
\mathbf{v}_{t+1} = \mathbf{v}^*, 
 \quad \mathcal{M}_{t+1} = \mathcal{M}_t+\mathcal{M}(\mathbf{v}^*, \mathcal{M}_t),
\label{eq:update}
\end{equation}
where $\mathcal{M}(\mathbf{v}^*, \mathcal{M}_t)$ means the new memory generated by SAM~\cite{ravi2024sam2} at the new view $\mathbf{v}^*$, and $\Delta \mathcal{A}(\mathbf{v}^*)$ is the change in the foreground Gaussian set after MFF based on the new mask obtained from SAM at the NBV $\mathbf{v}^*$. The coverage indicator $\tau_{t+1} = (\phi_t + \Delta \phi) / 180^{\circ}$ for a single direction virtual drone.

\section{Experiments}
\subsection{Experimental settings}
\begin{figure*}[t]
  \centering
    \setlength{\abovecaptionskip}{0.1cm}
  \includegraphics[width=1\linewidth]{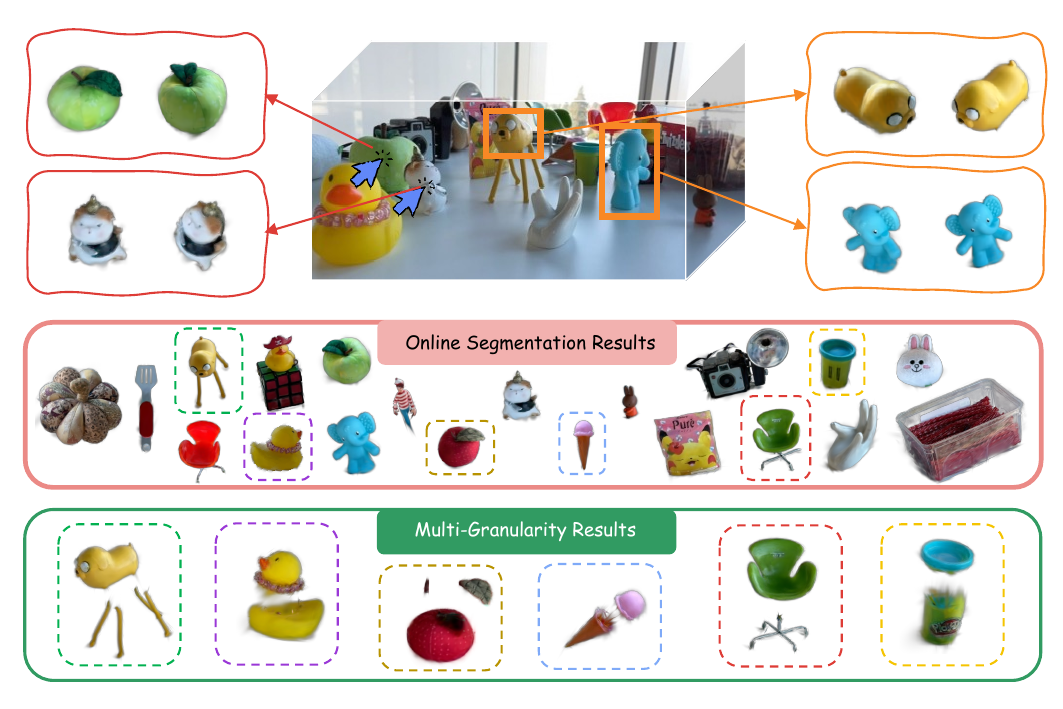}
  \caption{Online segmentation results of SAGO on the Figurines scene. The granularity is simply determined by the initial mask obtained via SAM prompted by the user, and the segmentation is iteratively refined through NBV-based online updates.}
  \label{fig:exp0}
\end{figure*}

\begin{figure*}[t]
  \centering
  \setlength{\abovecaptionskip}{0.cm}
  \includegraphics[width=1\linewidth]{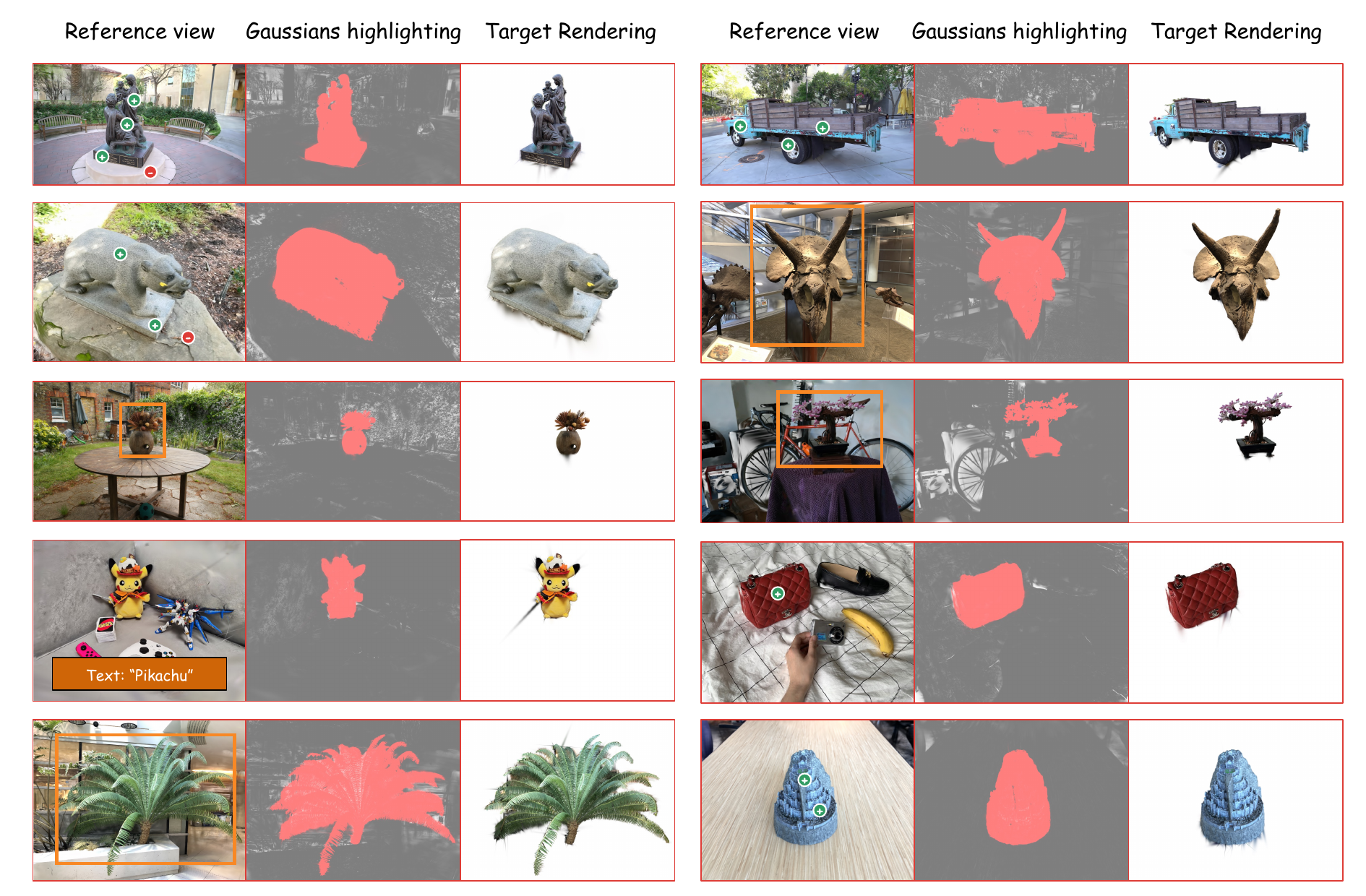}
  
  \caption{\textbf{Generalization test of SAGO.} We select 10 scenes from 5 datasets~\cite{mildenhall2019llff,kerr2023lerf, barron2022mip, knapitsch2017tanks, ren2022NOVS} and apply SAGO to segment the object described by the user via prompts like clicks, a box and text. The average segmentation latency per interaction is 500 ms.
  }
  \label{fig:generalization}
\end{figure*}

\textbf{Datasets.} To evaluate the effectiveness of our approach, we collect 3D scenes from multiple sources that encompass the majority of existing 3DGS segmentation datasets, including SPIn-NeRF~\cite{mirzaei2023spin}, NVOS~\cite{ren2022NOVS}, LLFF~\cite{mildenhall2019llff}, 3D-OVS~\cite{liu2023weakly}, LERF-Mask~\cite{kerr2023lerf}, MipNerf-360~\cite{barron2022mip}, T \& T~\cite{knapitsch2017tanks}, and PKU-DyMVHumans~\cite{zheng2024pku}. These datasets serve as the basis for our qualitative analysis. For quantitative comparison, we utilize SPIn-NeRF, NVOS, 3D-OVS, and LERF-Mask.

\noindent\textbf{Implementation Details.} 
We employ SAM2~\cite{ravi2024sam2} as our segmentation and tracking agent. The rendering resolution of each virtual drone viewpoint is set to $512 \times 512$, which matches the input resolution used by SAM2. The search space consists of yaw increments $\Delta\phi \in \{90^\circ, 60^\circ, 30^\circ\}$ and pitch angles $\theta \in \{\theta_0, 0^\circ, 30^\circ, 60^\circ\}$. We use Grounding-DINO~\cite{groundingDino} to transform the text prompt to a bounding box when using text prompts. All experiments are conducted on a single NVIDIA RTX 4090 GPU.

\subsection{Qualitative Results}
\label{sec:qual}
A key advantage of our method is that it does not need a time-consuming setup stage before performing online segmentation. We provide sufficient qualitative results in~\cref{fig:exp0} and~\cref{fig:generalization} to demonstrate this advantage and generalization capacity of our method across various scenes and datasets. 

\cref{fig:exp0} illustrates the online segmentation results of SAGO on the Figurines scene. The granularity of segmentation is determined by the initial mask obtained via SAM prompted by the user, and the segmentation is iteratively refined through NBV-based online updates. \cref{fig:generalization} further tests the generalization capacity of SAGO by applying it to ten scenes from five different datasets, performing online segmentation of the main objects without any scene-specific setup. The average segmentation latency per interaction is 500 ms, demonstrating the efficiency and adaptability of our method across diverse scenarios. 

\subsection{Quantitative Comparison}
\label{sec:quant}

\begin{table*}[h]
\centering
\renewcommand{\arraystretch}{1.5}
\setlength{\tabcolsep}{2.0mm}
\footnotesize
\begin{tabular}{l c cc cc cc}
\noalign{\hrule height 1.2pt}
\multirow{2}{*}{Method} & \multirow{2}{*}{Type} & \multicolumn{2}{c}{SPIn-NeRF} & \multicolumn{2}{c}{NVOS} & \multicolumn{2}{c}{Time} \\
\cline{3-4}\cline{5-6}\cline{7-8}
& & mIoU & mAcc & mIoU & mAcc & Setup & Seg. \\
\hline
MVSeg~\cite{mirzaei2023spin}  & Offline & 90.4 & 98.8 & - & - & - & - \\
NVOS~\cite{ren2022NOVS} & Offline & - & - & 70.1 & 92.0 & - & - \\
ISRF~\cite{goel2023interactive} & Offline & 71.5 & 95.5 & 83.8 & 96.4 & - & - \\
SA3D~\cite{cen2023SA3D} & Offline & 91.9 & 98.8 & 90.3 & 98.2 & 2\textasciitilde5m & 15\textasciitilde30s \\
LangSplat~\cite{qin2024langsplat} & Offline & 69.5 & 94.5 & 74.0 & 94.0 & \tiny$\sim$\footnotesize2.5h & - \\
SAGA~\cite{cen2025segment} & Offline & \cellcolor{red!25}\textbf{93.4} & \cellcolor{yellow!25}99.2 & \cellcolor{yellow!25}\underline{92.6} & \cellcolor{yellow!25}\underline{98.6} & \tiny$\sim$\footnotesize1h & 10ms \\
\hline
Flashsplat~\cite{shen2024flashsplat} & Online & - & - & 91.8 & \underline{98.6} & 30\textasciitilde60s & 1\textasciitilde2s \\
iSegMan~\cite{zhao2025isegman} & Online & 92.4 & \cellcolor{orange!25}99.1 & 92.0 & 98.4 & 50s\textasciitilde90s & 4\textasciitilde6s \\
GaussianCut~\cite{jain2024gaussiancut} & Online & \cellcolor{yellow!25}\underline{92.9} & \cellcolor{yellow!25}\underline{99.2} & \cellcolor{orange!25}92.5 & 98.4 & \textcolor{ForestGreen}{\textbf{N/A}} & 40\textasciitilde60s \\
\hline
Ours  & Online & \cellcolor{orange!25}92.5 & \cellcolor{red!25}\textbf{99.3} & \cellcolor{red!25}\textbf{92.7} & \cellcolor{red!25}\textbf{98.7} & \textcolor{ForestGreen}{\textbf{N/A}} & \textbf{0.4\textasciitilde0.9s} \\
\noalign{\hrule height 1.2pt}
\end{tabular}
\vspace{2mm}
\caption{Comparison of interactive 3D segmentation on SPIn-NeRF~\cite{mirzaei2023spin} and NVOS~\cite{ren2022NOVS}. \colorbox{red!25}{First}, \colorbox{orange!25}{second}, and \colorbox{yellow!25}{third} best results are highlighted. }
\label{tab:quant}
\end{table*}

\begin{figure}[t]
\centering
\begin{minipage}{0.5\textwidth}
\centering
\small
\setlength{\tabcolsep}{2mm}
\renewcommand{\arraystretch}{1.3}
\captionof{table}{\textbf{Quantitative comparison on LERF-Mask (mIoU $\uparrow$).}}
\label{tab:lerf_mask}
\resizebox{\linewidth}{!}{%
\begin{tabular}{lcccc}
\toprule
Method & Figurines & Ramen & Teatime & Average \\
\midrule
Lerf\cite{kerr2023lerf} & 33.5 &  28.3 &  49.7       &37.1        \\
Gau-Group\cite{GauGroup} & 69.7  & 77.0 & 71.7       & 66.1     \\

OmniSeg3D\cite{ying2024omniseg3d}    & 87.1  & 77.3 & 73.8      & 79.4      \\
ClickGaussian\cite{choi2024click}    & 93.2  &\textbf{90.9} & 83.2  &  89.1    \\
\hline
Ours        & \textbf{93.9}  & 90.6 & \textbf{88.6}  & \textbf{91.0}               \\   
\bottomrule
\end{tabular}
}
\end{minipage}%
\hfill
\begin{minipage}{0.47\textwidth}
\captionof{table}{\textbf{Quantitative comparison on 3D-OVS dataset (mIoU $\uparrow$).}}
\label{tab:3D-ovs}
\resizebox{\linewidth}{!}{%
\begin{tabular}{lcccccc}
  \toprule
    Method & bed & bench & room & sofa & lawn & mean\\
    \hline
    LERF~\cite{kerr2023lerf} & 73.5 & 53.2 & 46.6 & 27.0 & 73.7 & 54.8\\
    3D-OVS~\cite{liu2023weakly} & 89.5 & 89.3 & 92.8 & 74.0 & 88.2 & 86.8\\
    LangSplat~\cite{qin2024langsplat} & 92.5 & 94.2 & 94.1 & 90.0 & 96.1 & 93.4\\
    N2F2~\cite{bhalgat2024n2f2} & 93.8 & 92.6 & 93.5 & 92.1 & 96.3 & 93.9\\
    SAGA~\cite{cen2025segment} & 97.4 & 95.4 & \textbf{96.8} & 93.5 & \textbf{96.6} & 96.0\\
    \hline
    Ours & \textbf{98.5} & \textbf{96.6} & 95.7 & \textbf{94.3} & 95.9 & \textbf{96.2}\\
    \bottomrule
  \end{tabular}
}
\end{minipage}
\end{figure}

\begin{wrapfigure}{r}{0.5\textwidth}
\vspace{-10mm}
  \centering
  \includegraphics[width=0.42\textwidth]{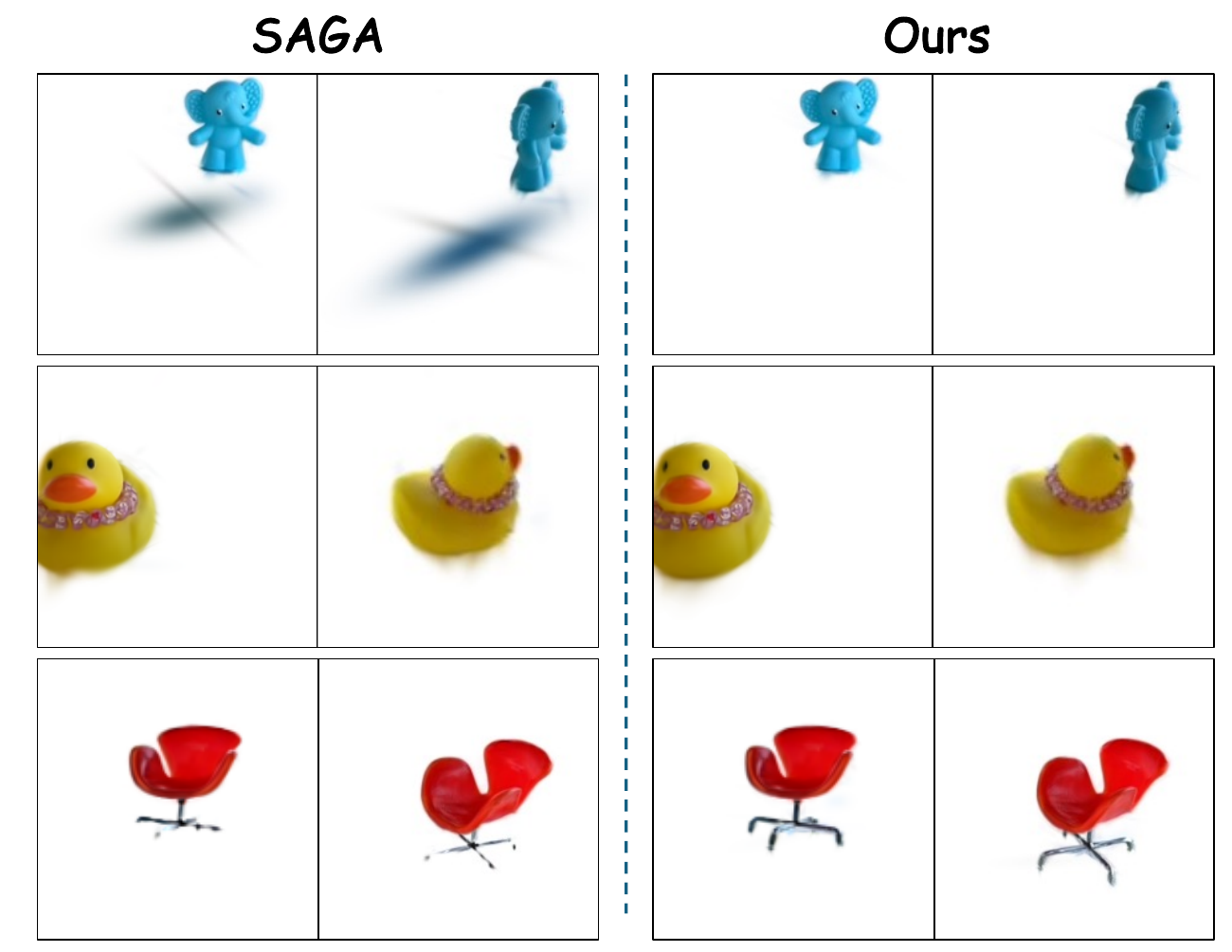}
  \caption{Comparison to SAGA~\cite{cen2025segment}, the milestone of offline methods.}
  \label{fig:saga}
  \vspace{-8mm}
\end{wrapfigure}

\vspace{-2mm}
We perform quantitative comparisons on four mainstream datasets (LERF-Mask~\cite{kerr2023lerf}, 3D-OVS~\cite{liu2023weakly}, SPIn-NeRF~\cite{mirzaei2023spin}, and NVOS~\cite{ren2022NOVS}), with additional efficiency analysis on NVOS and SPIn-NeRF. As shown in~\cref{tab:quant,tab:lerf_mask,tab:3D-ovs}, by comparing with current state-of-the-art (SOTA) methods, our approach achieves comparable—or even superior—performance in both mIoU and mAcc, without requiring any scene-specific setup.

\noindent\textbf{Results on SPIn-NeRF and NVOS.} As shown in~\cref{tab:quant}, on the NVOS dataset, our method achieves the best mIoU of 92.7\% and mAcc of 98.7\%, outperforming all compared methods including both offline and online approaches. Compared to the offline method SAGA~\cite{cen2025segment}, which requires approximately 1 hour of setup time for feature distillation, our method improves mIoU by 0.1\% and mAcc by 0.1\% on NVOS without any setup overhead. On SPIn-NeRF, our method achieves 92.5\% mIoU and the highest mAcc of 99.3\%, ranking second and first respectively, closely matching the best offline method SAGA (93.4\% mIoU). Compared to online methods, our approach surpasses iSegMan~\cite{zhao2025isegman} by 0.1\% mIoU and 0.2\% mAcc on SPIn-NeRF, and by 0.7\% mIoU and 0.3\% mAcc on NVOS, while being significantly faster. Notably, in contrast to the existing setup-free SOTA method GaussianCut~\cite{jain2024gaussiancut}, our method improves mIoU by 0.2\% and mAcc by 0.3\% on NVOS, and achieves comparable mIoU with 0.1\% higher mAcc on SPIn-NeRF, while reducing segmentation time from 40\textasciitilde60 seconds to 0.4\textasciitilde0.9 seconds, achieving a speedup of over \textbf{50$\times$}.

\noindent\textbf{Results on LERF-Mask.} As shown in~\cref{tab:lerf_mask}, our method achieves the highest average mIoU of 91.0\% across the three scenes, surpassing ClickGaussian~\cite{choi2024click} by 1.9\%. In particular, our method obtains the best performance on the Figurines (93.9\%) and Teatime (88.6\%) scenes, with significant improvements of 5.4\% over ClickGaussian on Teatime. On the Ramen scene, our method achieves a competitive 90.6\%, closely matching the best result of 90.9\%.

\noindent\textbf{Results on 3D-OVS.} As shown in~\cref{tab:3D-ovs}, our method achieves the highest mean mIoU of 96.2\%, outperforming the previous best method SAGA~\cite{cen2025segment} by 0.2\%. Specifically, our approach achieves the best results on three out of five scenes (\textit{bed}: 98.5\%, \textit{bench}: 96.6\%, \textit{sofa}: 94.3\%), and remains highly competitive on the remaining two scenes.

\subsection{3D Assets Extraction and Online Editing}

\begin{figure*}[t]
  \centering
  \setlength{\abovecaptionskip}{0.cm}
  \includegraphics[width=1\linewidth]{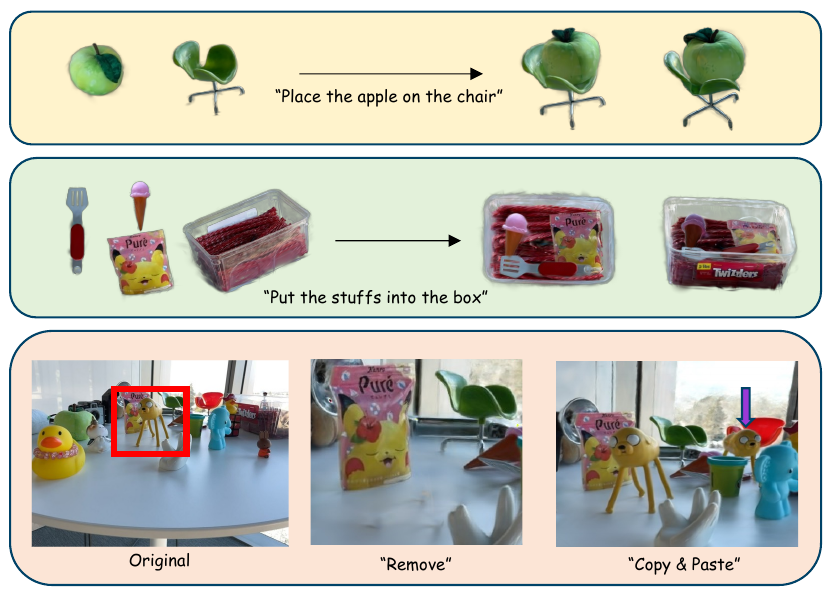}
  
  \caption{\textbf{Online Editing 3DGS via SAGO.} SAGO enables editing on a raw 3DGS scene in test-time.
  }
  \label{fig:editing}
\end{figure*}

The core advantage of our method is its ability to extract 3D assets and perform online editing. Our method extracts user-specified 3D assets from a raw 3DGS scene in the shortest time (less than 1 second), and after the extraction is completed, users can directly perform online editing on these 3D assets. As shown in~\cref{fig:editing}, we demonstrate the editing results of SAGO on the Figurines scene. After extracting the 3D asset of the leftmost figurine, we perform online editing operations such as relocation, removing, combination, copying on this asset, showcasing our method as a powerful tool for interactive 3D content creation and manipulation in 3DGS scenes.

\subsection{Ablation Study}
\label{sec:ablation}
\noindent\textbf{NBV strategy vs. Real views.} The motivation behind our virtual drone strategy is to explore the Next-Best-View (NBV) for online segmentation, while the real views are the actual camera views in the original 3DGS scene. 
To validate the effectiveness of our NBV strategy, we compare it against using only real views from the original scene. As shown in \cref{tab:Ablation}, removing the NBV strategy (NBV \ding{55}) causes a dramatic performance drop across all datasets, with LERF-Mask (complex and heavy occlusion) dropping by 54.3\% mIoU to 36.7\%, NVOS by 3.6\%, SPIn-NeRF by 2.4\%, and 3D-OVS by 4.6\%. This demonstrates that virtual drone viewpoints generated by our NBV strategy are crucial for exploring occluded regions and improving segmentation accuracy beyond what real views alone can provide. The more complex the scene and heavier the occlusion, the more significant the performance drop without NBV, highlighting the importance of our NBV strategy in challenging scenarios.

\noindent\textbf{Splat-based MFF vs. Center-based MFF.} Mask-shaped Frustum Filtering (MFF) filters out the background Gaussians based on the 2D mask at each state. We compare our center-based MFF with a splat-based MFF that retains Gaussians whose splats have any overlap with the 2D mask. As shown in \cref{tab:Ablation}, removing the center-based MFF (Center-based MFF \ding{55}) results in a performance drop across all datasets, with LERF-Mask dropping by 0.6\% mIoU, NVOS by 0.2\%, SPIn-NeRF by 1.4\%, and 3D-OVS by 0.2\%. Although the performance drop is not significant, the center-based MFF is computationally simpler, which is why we adopt it in our implementation.


  

\begin{table}[t]
\centering
\renewcommand{\arraystretch}{1.8}
\setlength{\tabcolsep}{0.4mm}
\tiny
\resizebox{0.85\linewidth}{!}{%
\begin{tabular}{cccccc}
\noalign{\hrule height 1.2pt}
\multirow{2}{*}{NBV} & \multirow{2}{*}{\makecell{Center-based\\MFF}} & \multicolumn{4}{c}{mIOU} \\
\cline{3-6}
& & LERF-Mask & NVOS & SPIn-NeRF & 3D-OVS\\
\hline
\ding{51} & \ding{51} & \textbf{91.0} & \textbf{92.7} & \textbf{92.5} & \textbf{96.2} \\
\ding{55} & \ding{51} & 36.7\,(\textcolor{red}{\tiny-54.3$\downarrow$}) & 89.1\,(\textcolor{red}{\tiny-3.6$\downarrow$}) & 90.1\,(\textcolor{red}{\tiny-2.4$\downarrow$}) & 91.6\,(\textcolor{red}{\tiny-4.6$\downarrow$}) \\
\ding{51} & \ding{55} & 90.4\,(\textcolor{red}{\tiny-0.6$\downarrow$}) & 92.5\,(\textcolor{red}{\tiny-0.2$\downarrow$}) & 91.1\,(\textcolor{red}{\tiny-1.4$\downarrow$}) & 96.0\,(\textcolor{red}{\tiny-0.2$\downarrow$}) \\
\noalign{\hrule height 1.2pt}
\end{tabular}
}%
\vspace{2mm}
\caption{Ablation study on multiple datasets. }
\label{tab:Ablation}
\end{table}

\section{Conclusion}
In this paper, we present SAGO, a novel setup-free framework for interactive segmentation of 3D Gaussians. Central to our approach is the introduction of \textbf{virtual drones}, a key innovation that reformulates the 3D segmentation problem as a multi-step online Next-Best-View (NBV) planning task. By leveraging virtual drones and NBV planning, SAGO effectively eliminates the time-consuming per-scene setup required by existing methods, while maintaining practical interaction times (typically within 1 second).

\noindent\textbf{Limitations}. (1) The center-based MFF takes each Gaussian's center as a representative point. Since Gaussians typically have volume and cannot be treated as ideal points, SAGO currently cannot address jagged boundaries. However, this does not pose a barrier to subsequent work based on SAGO, as online boundary trimming methods such as GaussianTrimmer~\cite{liao2026gaussiantrimmer} can be introduced to improve boundary segmentation performance. (2) SAGO requires a high-quality initial view for effective state initialization, which may restrict its applicability when the initial view is suboptimal or unavailable. Future work could investigate strategies for using v-drones to actively explore an optimal initial view.


\section*{Acknowledgements}
This work is financially supported by Guangdong Provincial Key Laboratory of Ultra High Definition Immersive Media Technology (Grant No. 2024B1212010006). This work is also financially supported for Fundamental and Interdisciplinary Disciplines Breakthrough Plan of the Ministry of Education of China (Grant No. JYB2025XDXM413), Outstanding Talents Training Fund in Shenzhen, Shenzhen Science and Technology Program (Grant No. SYSPG20241211173440004).

%
%
\bibliographystyle{splncs04}
\bibliography{main}

\clearpage
\setcounter{page}{1}
\maketitlesupplementary

\section{Computational analysis}
To further analyze the efficiency of SAGO, we conducted a statistical analysis of the segmentation time and the number of segmentation steps for four datasets including SPIn-NeRF~\cite{mirzaei2023spin}, NVOS~\cite{ren2022NOVS}, 3D-OVS~\cite{liu2023weakly}, and LERF-Mask~\cite{kerr2023lerf}.
\begin{figure*}[h]
  \centering
  \vspace{-0.5cm}
    \setlength{\abovecaptionskip}{0.3cm}
    \setlength{\belowcaptionskip}{-0.2cm}
  \includegraphics[width=1\linewidth]{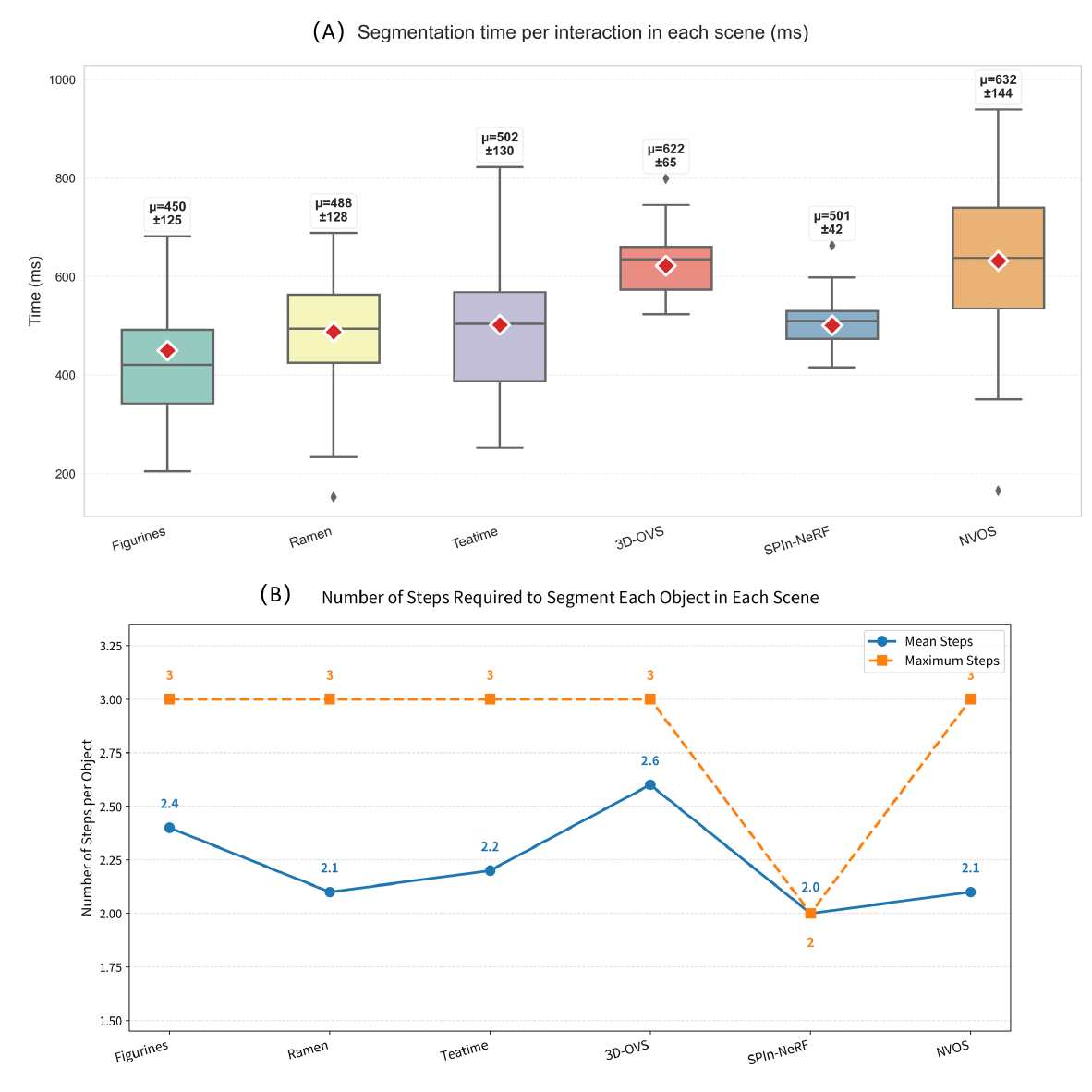}
  \caption{\textbf{Computational analysis.}(A) Segmentation time per interaction in each scene. (B) Average and maximum step per segmentation. Note that `Figurines', `Ramen' and `Teatime' are from LERF-Mask dataset. Due to the complexity of these three scenes, we provide a separate analysis.
  }
  \label{fig:time}
\end{figure*}

\subsection{Inference Time Analysis}
As shown in~\cref{fig:time} (A), our statistical analysis across multiple datasets indicates that the segmentation time per interaction ranges from 0.3 to 1 second, with an average of approximately 0.5 seconds. The results demonstrate that even without the setup stage, our proposed method exhibits strong stability in terms of interactive segmentation latency.
The segmentation time we report does not include the initial mask time generated during user interaction, which typically takes 0.2-0.5 seconds. We consider this time as part of the interaction time, and we also exclude this time when comparing with other methods.

\subsection{Markov State Transfer Count Analysis}
As described in~\cref{sec:NBV}, in the worst case the yaw step size is limited to $30^\circ$, requiring up to 6 Markov states ($180^\circ/30^\circ = 6$) to achieve the full angular coverage needed for segmentation. Fortunately, as shown in~\cref{fig:time} (B), thanks to the carefully designed NBV planning strategy, experiments on four datasets~\cite{mirzaei2023spin, ren2022NOVS, liu2023weakly, kerr2023lerf} show that no scene required more than three Markov states to complete the segmentation.

\begin{wrapfigure}{r}{0.5\textwidth}
\vspace{-1cm}
  \centering
  \includegraphics[width=0.45\textwidth]{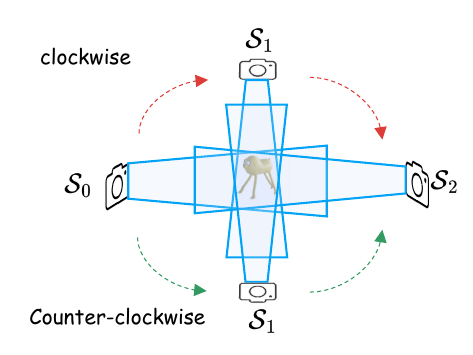}
  \caption{\textbf{Theoretical minimum state count.} We use bidirectional NBV planning, so the theoretical minimum number of Markov states is \textbf{2} (i.e., $S_1$ and $S_2$).}
  \label{fig:minimum}
  \vspace{-4mm}
\end{wrapfigure}

\noindent\textbf{Theoretical minimum state count.} When counting the number of states, we exclude the initial state \(S_0\). As shown in~\cref{fig:minimum}, our bidirectional NBV implies that the theoretical minimum number of Markov states is \textbf{2} (\(S_1\) and \(S_2\)). In practice, our experiments show that most scenes can be fully segmented using only 2--3 Markov states, demonstrating the effectiveness of our NBV planning strategy. Our NBV planning for the virtual drone prioritizes paths with the minimum state count. It only explores paths with smaller yaw spans when an excessively large span causes the rendered texture of the target object to become overly large, thereby degrading SAM2's tracking performance. In addition to relying on SAM2 tracking, our method for acquiring new-view masks also incorporates the 3D position of the target object as an additional spatial constraint to assist segmentation. This is one of the reasons why our method can complete segmentation in most scenes using only 2--3 Markov states.


\section{Implementation details}
\textbf{mIOU threshold}. In our implementation, we set the $\sigma$ threshold in~\cref{eq:nbv_planning} to 0.95. Our guiding principle is that segmentation from a new viewpoint must not degrade the rendering quality of the initial viewpoint. We use~\cref{eq:nbv_planning} to identify a suitable viewpoint at which SAM2 can track the target effectively. If the yaw span is too small, more Markov states are required, which significantly increases the segmentation time. Therefore, our proposed method first explores viewpoints with a large yaw span (90°). Only if the tracking performance is poor (first-view mIoU < $\sigma$ ) does it then explore smaller yaw spans. As shown in~\cref{fig:time}, the maximum number of steps required for segmentation is 3, which demonstrates the efficiency of our NBV planning strategy.


\noindent\textbf{Mask Expansion Coefficient for MFF.} In the implementation of Mask-shaped Frustum Filtering (MFF), we apply a mask expansion operation to the segmentation mask obtained from SAM2 before performing the filtering. We apply a mask expansion operation with a mask expansion coefficient $e$ to the binary mask obtained from SAM2 before performing MFF. In our experiments, we set the mask expansion coefficient $e$ to 3.

\begin{figure*}[h]
  \centering
  \vspace{-0.5cm}
    \setlength{\abovecaptionskip}{0.3cm}
    \setlength{\belowcaptionskip}{-0.2cm}
  \includegraphics[width=1\linewidth]{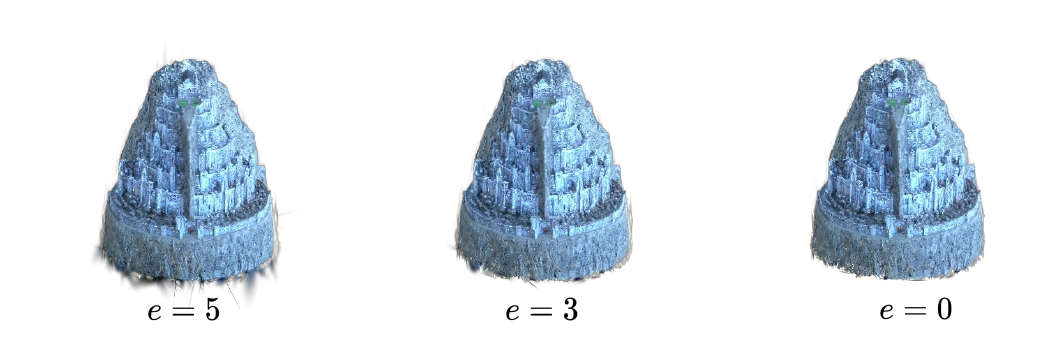}
  \caption{\textbf{Ablation study on Mask Expansion.} The three mask expansion coefficients (0, 3, and 5) are applied to the segmentation mask obtained from SAM2 before performing MFF.
  }
  \label{fig:mask_expansion}
\end{figure*}

As shown in~\cref{fig:mask_expansion}, we conducted an ablation study on the mask expansion coefficient for MFF. We tested three different coefficients (0, 3, and 5) applied to the segmentation mask obtained from SAM2 before performing MFF. From a practical perspective, there is no optimal mask expansion coefficient. This serves as a user option: if you want a more complete segmentation result that may have edge artifacts, you can choose a larger mask expansion coefficient; if you want sharper edges but may lose edge details, you can choose a smaller mask expansion coefficient.

\subsection{Anti-occlusion Mechanism}
\begin{figure*}[h]
  \centering
  \vspace{-0.5cm}
    \setlength{\abovecaptionskip}{0.3cm}
    \setlength{\belowcaptionskip}{-0.2cm}
  \includegraphics[width=0.8\linewidth]{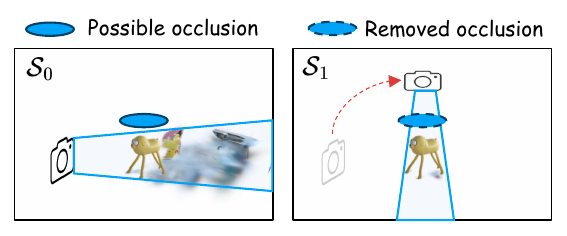}
  \caption{\textbf{Illustration of anti-occlusion mechanism.} Virtual drones can actively eliminate potential occlusions through the Markov process of online NBV planning. 
  }
  \label{fig:anti-occlusion}
\end{figure*}

In 3D scenes, mutual occlusions inevitably exist between multiple viewpoints. Prior works have devoted substantial effort to handling occlusions. However, our method can actively eliminate potential occlusions through the Markov process of online NBV planning. This gives our approach a significant advantage in dealing with complex scenes.
As shown in~\cref{fig:anti-occlusion}, at time \(t\), our virtual drone can use MFF to clear occlusions for the viewpoint at time \(t+1\), ensuring that the viewpoint at time \(t+1\) can fully cover the target object, thereby achieving complete segmentation.

\subsection{GUI-based Implementation for SAGO}

\begin{figure*}[h]
  \centering
  \vspace{-0.5cm}
    \setlength{\abovecaptionskip}{0.3cm}
    \setlength{\belowcaptionskip}{-0.2cm}
  \includegraphics[width=1\linewidth]{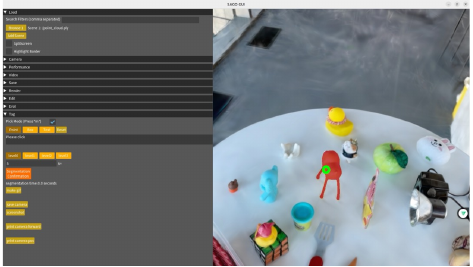}
  \caption{\textbf{Illustration of GUI-based implementation for SAGO.} The GUI allows users to interactively perform segmentation in 3D scenes, select target objects, visualize segmentation results, and adjust parameters such as the mask expansion coefficient for MFF.}
  \label{fig:gui}
\end{figure*}

\noindent To facilitate the practical application of SAGO, we have developed a GUI-based implementation that allows users to interactively perform segmentation in 3D scenes. The GUI provides an intuitive interface for users to select target objects, visualize segmentation results, and adjust parameters such as the mask expansion coefficient for MFF. We design this GUI based on SplatViz~\cite{barthel2024splatviz}.

\section{Additional Qualitative Results}
\subsection{Online Free Granularity Control}
The segmentation granularity of the interactive 3D segmentation results in our method can be completely determined by the 2D segmentation granularity. Users only need to select their preferred 2D segmentation granularity through simple interactive operations such as clicking or drawing bounding boxes, and this 2D granularity will be directly reflected in the 3D segmentation results.

\begin{figure*}[h]
  \centering
  \vspace{-0.5cm}
    \setlength{\abovecaptionskip}{0.3cm}
    \setlength{\belowcaptionskip}{-0.2cm}
  \includegraphics[width=1\linewidth]{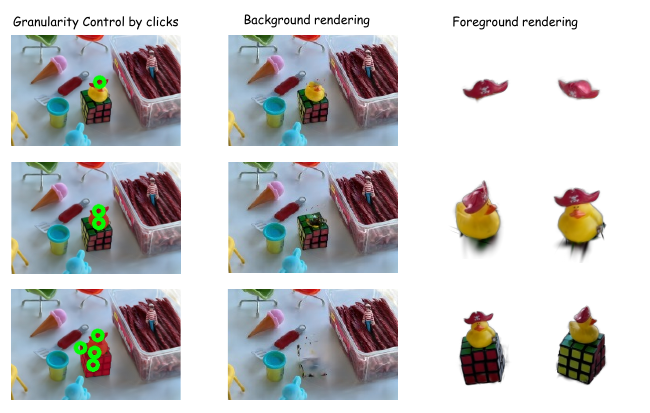}
  \caption{\textbf{Illustration of online free granularity control.} The 2D granularity selected by the user is directly reflected in the 3D segmentation results.}
  \label{fig:granularity}
\end{figure*}
As shown in~\cref{fig:granularity}, we demonstrate the online free granularity control capability of our method. By adjusting the 2D segmentation granularity through user interactions, users can achieve their desired level of detail in the 3D segmentation results, providing a flexible and user-friendly experience.

\begin{figure*}[h]
  \centering
  \vspace{-0.5cm}
    \setlength{\abovecaptionskip}{0.3cm}
    \setlength{\belowcaptionskip}{-0.2cm}
  \includegraphics[width=1\linewidth]{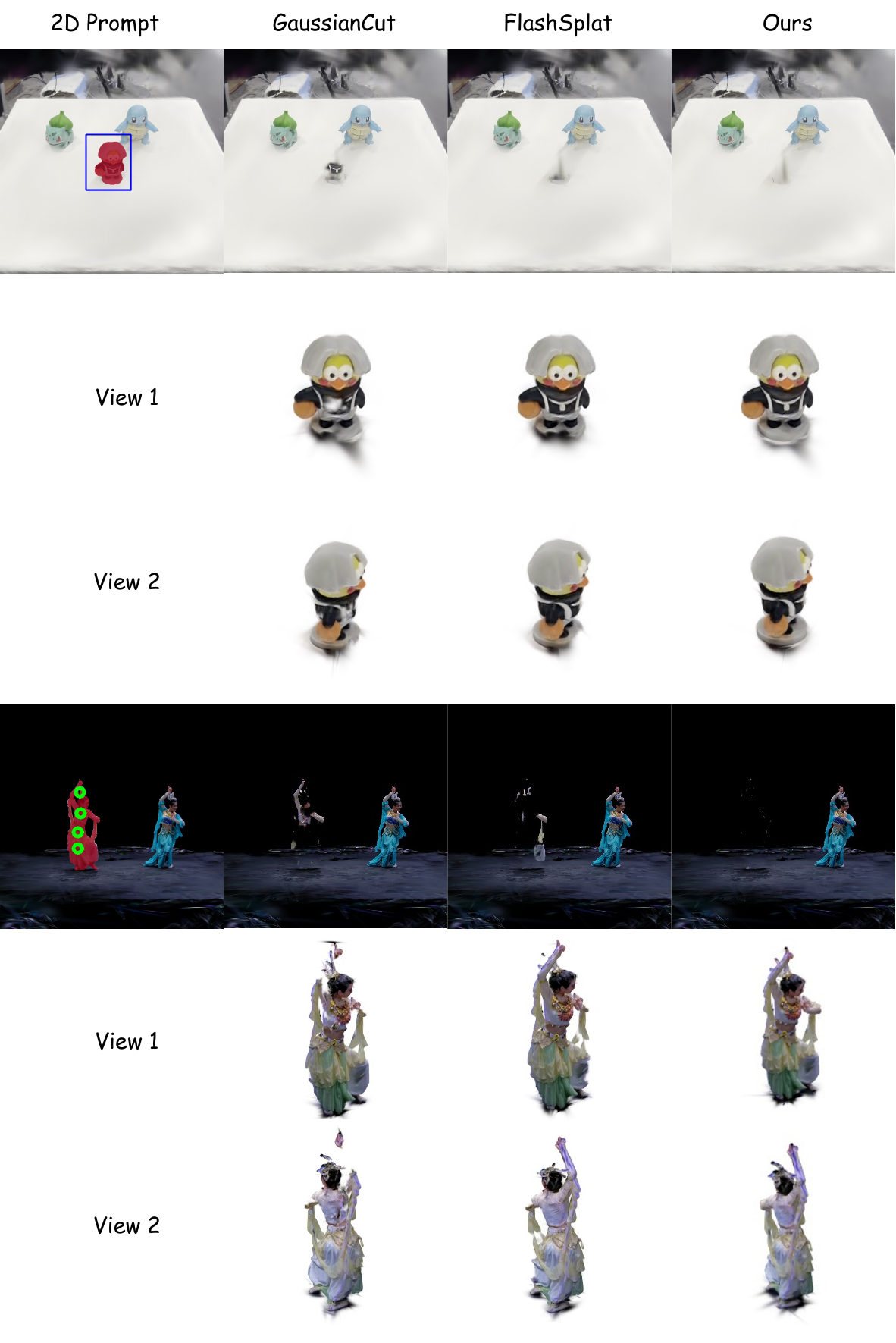}
  \caption{\textbf{Qaulitative comparison with other online methods~\cite{jain2024gaussiancut,shen2024flashsplat}.} Our method achieves superior segmentation quality and efficiency compared to other online methods, demonstrating its effectiveness in interactive 3D segmentation tasks.}
  \label{fig:comparison}
\end{figure*}
\subsection{Comparison with Other Online Methods}
\label{sec:comparison}


We conduct a qualitative comparison of our method with other online methods for interactive 3D segmentation.
As shown in~\cref{fig:comparison}, our method achieves superior segmentation quality and efficiency compared to other online methods, demonstrating its effectiveness in interactive 3D segmentation tasks.

\clearpage
\section{Limitations and Failure Cases}
\textbf{Dependence on good initial segmentation}. Our method relies on the initial segmentation results obtained from user interactions. If the initial segmentation is of poor quality, it may lead to suboptimal segmentation results in subsequent steps. Therefore, our method requires a high-quality initial interactive segmentation result to ensure the effectiveness of the overall segmentation process.

\noindent\textbf{No boundary trimming.} 
The current implementation of our method does not include a boundary trimming mechanism, which may lead to boundary artifacts in the segmentation results. This is because our Mask-shaped Frustum Filtering (MFF) operation is center-based, and since Gaussians have volume, especially at the edges where they often span across two objects, a purely center-based approach cannot effectively address edge artifacts. However, we are currently exploring the integration of a post-processing tool called GaussianTrimmer~\cite{liao2026gaussiantrimmer} to mitigate edge artifacts and improve the overall quality of the segmentation results.

\end{document}